\documentclass[sigconf]{acmart}
\AtBeginDocument{%
  \providecommand\BibTeX{{%
    \normalfont B\kern-0.5em{\scshape i\kern-0.25em b}\kern-0.8em\TeX}}}

\usepackage{graphicx}
\usepackage{subfigure}
\usepackage{amsmath}
\usepackage{amsfonts}
\usepackage{multirow}
\usepackage{comment}
\usepackage{hyperref}

\let\emptyset\varnothing

\newcommand{\Modelshortsp}{Hy-Transformer }
\newcommand{\Modelshort}{Hy-Transformer}

\newcommand{\Baselineshortsp}{STARE }
\newcommand{\Baselineshort}{STARE}

\usepackage{xcolor}
\definecolor{donghan}{RGB}{243, 101, 2}
\definecolor{Red}{RGB}{255, 3, 13}
\definecolor{Plum}{RGB}{142, 69, 133}

\setcopyright{acmcopyright}
\copyrightyear{2018}
\acmYear{2018}
\acmDOI{10.1145/1122445.1122456}

\acmConference[Woodstock '18]{Woodstock '18: ACM Symposium on Neural
  Gaze Detection}{June 03--05, 2018}{Woodstock, NY}
\acmBooktitle{Woodstock '18: ACM Symposium on Neural Gaze Detection,
  June 03--05, 2018, Woodstock, NY}
\acmPrice{15.00}
\acmISBN{978-1-4503-XXXX-X/18/06}

\begin{document}



\title{Improving  Hyper-Relational Knowledge Graph Completion}


\author{Donghan Yu}
\affiliation{%
  \institution{Carnegie Mellon University}
  \city{Pittsburgh}
  \country{USA}
}
\email{dyu2@cs.cmu.edu}

\author{Yiming Yang}
\affiliation{%
  \institution{Carnegie Mellon University}
  \city{Pittsburgh}
  \country{USA}
}
\email{yiming@cs.cmu.edu}

\renewcommand{\shortauthors}{Yu et al.}

\begin{abstract}
  Different from traditional knowledge graphs (KGs) where facts are represented as entity-relation-entity triplets, hyper-relational KGs (HKGs) allow triplets to be associated with additional relation-entity pairs (a.k.a qualifiers) to convey more complex information.
  How to effectively and efficiently model the triplet-qualifier relationship for prediction tasks such as HKG completion is an open challenge for research. This paper proposes to improve the best-performing method in HKG completion, namely STARE, by introducing two novel revisions:
  (1) Replacing the computation-heavy graph neural network module with light-weight entity/relation embedding processing techniques for efficiency improvement without sacrificing effectiveness; 
  (2) Adding a qualifier-oriented auxiliary training task for boosting the prediction power of our approach on HKG completion.
  The proposed approach consistently outperforms STARE  in our experiments on three benchmark datasets, with  significantly improved computational efficiency.
\end{abstract}





\maketitle

\section{Introduction}

Knowledge graphs (KGs) have received increasing attention in recent machine learning community due to their broad range of important applications including natural language processing \cite{dialogue,lin2019kagnet}, recommender systems \cite{kgrecsurvey}, computer vision \cite{chen2020knowledge}, and more. KGs typically sore factual information in the form of (head entity, relation, tail entity) triplet, such as (\textit{Alan Turing}, \textit{educated at}, \textit{Princeton University}).

Hyper-relational KGs (HKGs) go beyond conventional KGs by representing facts with more complex semantic information, e.g., using relation-entity pairs as the \textit{qualifiers} of triplets.
The combination of a triplet and its  qualifiers together is called a \textit{statement}. For example, one statement can be (\textit{Alan Turing}, \textit{educated at}, \textit{Princeton University}, (\textit{academic degree}, \textit{Doctorate}), (\textit{academic major}, \textit{Mathematics})) where (\textit{academic degree}, \textit{Doctorate}) and (\textit{academic major}, \textit{Mathematics}) are qualifiers. This statement conveys the information that Alan Turing studied at Princeton University and received a doctorate degree major in math from there. Such hyper-relational data is ubiquitous in KGs. For example in Freebase \cite{bollacker2008freebase}, more than 30\% of its entities are involved in such hyper-relational facts \cite{jf17k}.

KG completion task, which aims to predict 
semantically valid but unobserved triplets
based on the observed ones, is an important task and has been intensively studied for triplet-based KGs in recent years \cite{kgsurvey}. However, this task remains underexplored for hyper-relational KGs. 
Existing works often lose semantic information in statement representation learning 
as a results of ignoring the paired relationship between qualifier entity and qualifier relation
\cite{jf17k,rae,hype,liu2020generalizing}, not differentiating triplets from qualifiers \cite{NaLP}, or failing to model the interaction among multiple qualifiers \cite{hinge}. 
The most successful approach so far is STARE \cite{stare}, which uses
a graph neural network (GNN) module to 
improve entity and relation embeddings and a Transformer \cite{transformer} encoder network 
to model the interactions among 
qualifiers and the triplet being qualified
in one statement. Although STARE achieved the state-of-the-art HKG completion performance, its GNN module causes a large computation overhead, which would limit its practical success to very large applications.

In this paper we aim to address the computational efficiency issue in STARE as well as to improve its prediction performance, with the following strategies:
(1) Replacing the computation-heavy GNN module with light-weight entity/relation embedding processing techniques; 
  (2) Adding an qualifier-oriented auxiliary training task for boosting the prediction power on HKG completion.
Experiment results on three benchmark datasets show that our model achieves the new state-of-the-art results and is significantly more efficient than \Baselineshort.

\section{Method}
\label{method}

We begin by introducing mathematical notations. In a hyper-relational KG $\mathcal{G}_H$, we denote the set of entities and relations as $\mathcal{V}$ and $\mathcal{R}$ respectively. The total number of entities is $N$ and the number of relations is $M$. The edge connecting them, which we call a statement (or fact), is expressed in the domain $\mathcal{V} \times \mathcal{R} \times \mathcal{V} \times \mathcal{P}(\mathcal{R} \times \mathcal{V})$ where $\mathcal{P}$ denotes the power set. It's usually written as $(m_h, m_r, m_t, \mathcal{Q})$ where $(m_h, m_r, m_t)$ is the \textit{main triplet} of the statement containing head entity $m_h \in \mathcal{V}$, relation $m_r \in \mathcal{R}$ and tail entity $m_t \in \mathcal{V}$ respectively. $\mathcal{Q}$ is the set of \textit{qualifiers} consisting $n$ relation-entity pairs $\{(q_{ri}, q_{ei})\}_{i=1}^n$ where $q_{ri} \in \mathcal{R}$ and $q_{ei} \in \mathcal{V}$. Note that the number of qualifier pairs $n$ can be different for different statements. $n$ can also be $0$ when one statement only has the main triplet ($\mathcal{Q}=\emptyset$ is an empty set in this case). 


The completion task on hyper-relational KGs is that, following previous setting \cite{hinge,stare}, given an incomplete statement $(m_h, m_r, ?, \mathcal{Q})$ or $(?, m_r, m_t, \mathcal{Q})$ where its head or tail entity of the main triplet is missing, the model is required to predict the missing entity from entity set $\mathcal{V}$. 

\subsection{Recap of STARE}
\label{recap}

We first introduce the base of our model: STARE\footnote{The original paper proposes different variants of models. Our model is based on STARE+MskTrf, which we refer to STARE for simplicity.}. Similar to the masking mechanism in BERT \cite{devlin2018bert}, \Baselineshortsp place a special [MASK] entity in the position of unknown entity in the input statement. For example, if the tail entity is unknown, the input is $(m_h, m_r, \text{[MASK]}, \mathcal{Q})$. Then it's flattened into a sequence $s  = [m_h, m_r, \text{[MASK]},$ $ q_{r1}, q_{e1},$ $\cdots, q_{rn}, q_{en}]$ following the order from main triplet to qualifiers. The maximum input length is set as $T$ and the sequence will be padded if its length is less than $T$.

Suppose the initial entity embedding matrix is $E\in \mathbb{R}^{(N+1)\times d}$ where the [MASK] entity takes the last row of embedding matrix and $d$ is the embedding dimension. The initial relation embedding matrix is  $R \in \mathbb{R}^{M\times d}$. STARE first updates the embedding matrices with a graph neural network (GNN) module:
\begin{align}\label{eq:embed}
    \hat{E}, \hat{R} & = \text{GNN}(E, R, \mathcal{G}_H)
\end{align}
where $\hat{E} \in \mathbb{R}^{(N+1)\times d}, \hat{R} \in \mathbb{R}^{M\times d}$ are the updated entity and relation embeddings respectively. By doing this, the local context information of entities has been explicitly encoded into their representations. Then based on the input sequence $s$, STARE queries the entity and relation embeddings from $\hat{E}$ and $\hat{R}$ and concatenate them to form the statement representation $S$: 
\begin{align}\label{query}
    S = [\hat{E}_{m_h}; \hat{R}_{m_r}; \hat{E}_{\text{mask}}; \cdots; \hat{R}_{q_{ri}}; \hat{E}_{q_{ei}}; \cdots] \in \mathbb{R}^{T\times d}
\end{align}
Then $S$ is 
forwarded to a Transformer encoder module and the output is the updated statement representation $\hat{S}$:
\begin{align}\label{eq:trans}
    \hat{S} & = \text{Transformer}(S) \in \mathbb{R}^{T\times d}
\end{align}
where, for simplicity, we assume the dimension of hidden representation in the Transformer model is also $d$. Then the representation of [MASK] entity is selected to pass through a feed-forward neural network $f: \mathbb{R}^d \rightarrow \mathbb{R}^d$ and dot product with each entity embedding to calculate the similarity values. Finally an element-wise sigmoid function is applied on the values to output the probability of each entity being the missing entity:
\begin{align}\label{eq:sig}
    o & = \text{Sigmoid}(f(\hat{S}_{\text{ind}_{\text{mask}}}) \hat{E}_{0:N-1}^T) \in \mathbb{R}^{1\times N}
\end{align}
where $o_i$ is the probability entity $i$ being the missing entity. $\text{ind}_{\text{mask}}$ is the index of the [MASK] entity in the input sequence. $\hat{E}_{0:N-1}$ is the processed entity embedding matrix without the embedding of [MASK] entity, which should be avoided as final output. 
Next, we'll introduce the two revisions we made on STARE.

\begin{figure}
\centering
\includegraphics[width=8.5cm]{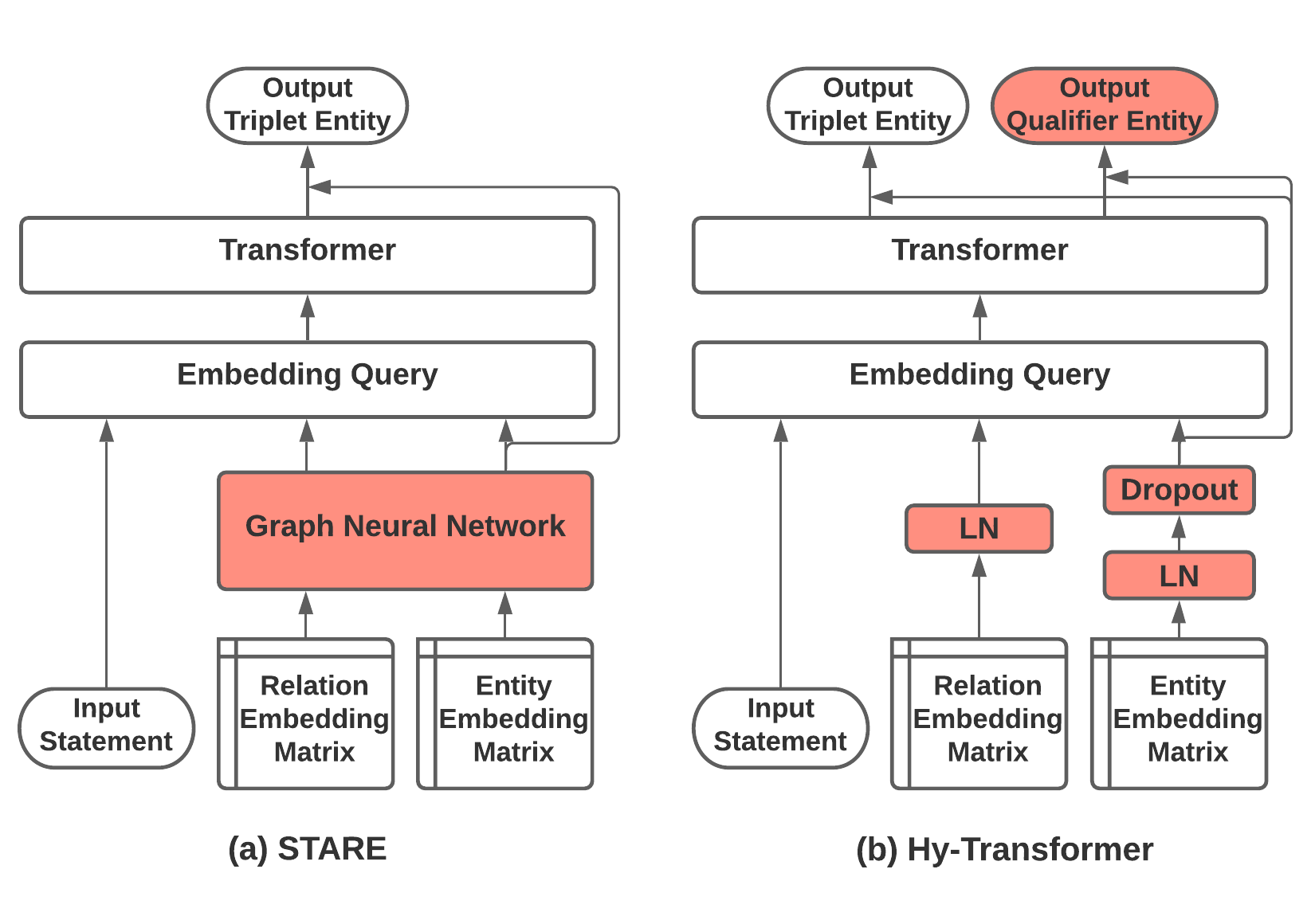}
\caption{A simple realization of our proposed model Hy-Transformer compared to STARE. The different parts are marked in red. LN is short for layer normalization \cite{layernorm}.}
\label{fig:model}
\end{figure}

\subsection{Replacing the GNN Module}
\label{replace}

We first propose to replace the GNN module for reducing computation cost. We show the inefficiency of GNN model by computing its time complexity of each training step: For each statement $(m_h, m_r, m_t, \mathcal{Q})$, the GNN module computes the combined embedding $h_{\mathcal{Q}}$ of qualifiers $\mathcal{Q}$ as follows and update the embeddings of entity $m_h$ and $m_t$ with it:
\begin{align}
    h_{\mathcal{Q}} = W \sum_{(q_{ri}, q_{ei}) \in Q} \phi(\hat{R}_{q_{ri}}, \hat{E}_{q_{ei}})
\end{align}
where $\phi: \mathbb{R}^d \times \mathbb{R}^d \rightarrow \mathbb{R}^d$ is a function to combine qualifier entity embedding and relation embedding and $W \in \mathbb{R}^{d \times d}$ is a projection matrix. The computation of $h_{\mathcal{Q}}$ involves matrix-vector multiplication which takes $O(d^2)$. Even if we ignore the time complexity of other operations, the total complexity is already $O(L_gZd^2)$ where $Z$ is the total number of statements in the HKG and $L_g$ is the number of GNN layers. This can be a large computation overhead since $Z$ can be very large. Note that this complexity can not be reduced by only updating the entities involved in the mini-batch of current training step, since the whole entity embedding matrix $\hat{E}$ is used in Equation (\ref{eq:sig}) for output entity prediction.

Another motivation to remove the GNN module is that intuitively the input statement already contains sufficient information to predict the unknown entity and the Transformer module is expressive enough to capture such information. Thus the encoding of local context of entities by graph convolution network might not be necessary. 

To replace the GNN module, we propose to use layer normalization \cite{layernorm} (LN) and dropout \cite{srivastava2014dropout} for entity and relation embedding processing:
\begin{align}
    \hat{E} & = \text{Dropout}(\text{LN}(E)) \in \mathbb{R}^{(N+1)\times d} \\
    \hat{R} & = \text{LN}(R) \in \mathbb{R}^{M\times d} 
\end{align}
where LN can introduce inductive bias about the embedding distribution by re-centering and re-scaling. Dropout has been empirically proved to be a effective regularization techniques when training large-scale neural network. This replacement reduces the complexity from $O(L_gZd^2)$ to $O(Nd + Md)$ which is no longer dependent to $Z$ and only linear to the dimension number $d$ instead of quadratic. In the Section \ref{main}, we'll also illustrate the efficiency improvement by empirical results on benchmark datasets. 

\subsection{Auxiliary Training Task}
\label{aux}

We further propose an auxiliary training task to boost the model performance on HKG completion: predicting the missing qualifier entity in a statement. For each training statement with qualifiers, and for each $i\in\{1,2,\cdots,n\}$, we construct an incomplete statement $(m_h, m_r, m_t,$ $\{(q_{r1}, q_{e1}), \cdots,$ $(q_{ri}, ?),$ $\cdots, (q_{rn}, q_{en})\})$ where the qualifier entity $q_{ei}$ is missing instead of triplet entity. Given the incomplete statement, our model needs to predict the missing qualifier entity. Similar to predicting missing entity in the main triplet introduced in Section \ref{recap}, we place a [MASK] entity in the position of missing qualifier entity and flat it into a sequence $s  = [m_h, m_r, m_t,$ $q_{r1}, q_{e1},$ $ \cdots q_{ri}, \text{[MASK]},$ $\cdots, q_{rn}, q_{en}]$. Then this sequence is input into our model and the representation of [MASK] entity after Transformer module is used for final entity prediction. The final training data is simply the union of the training data on the primary task and on the auxiliary task.

Although the primary task is to predict the entity of main triplet, we think this auxiliary training task can augment the training data to learn better interaction among the entities and relations in a statement. The effectiveness of this training strategy will be empirically demonstrated in Section \ref{qualifier}. 

Our final model is named as \textit{Hy-Transformer} where \textit{Hy} is the abbreviation of Hyper-relational KG. Figure \ref{fig:model} shows the comparison between our model and \Baselineshort.


\section{Experiment}

In this section, we demonstrate both the effectiveness and efficiency of our model by comprehensive experiments on three benchmark datasets. 

\subsection{Basic Setting}
\label{basic}

We conduct experiments on the following benchmark datasets: JF17K \cite{jf17k}, Wikipeople \cite{NaLP} and WD50K \cite{stare}, where the dataset statistics is shown in Table \ref{tab:data}. We follow the conventional train/valid/test split setting \cite{jf17k,NaLP,stare}. Following previous work \cite{conve,stare}, we train the model in 1-N setting using binary cross entropy loss with label smoothing. The optimizer is Adam \cite{adam} and the learning rate is 0.0001. The number of training epochs is 400 for JF17K and WD50K and 500 for Wikipeople. For the transformer module, the number of layer is 2 and the hidden dimension is 512 with dropout rate as 0.1. The dimension of initial entity/relation embedding is 200. The dropout rate for entity embedding matrix is 0.3. We perform grid search to find the best performing hyperparameters similar in \cite{stare}. All the experiments are conducted on one Nvidia 2080-Ti GPU.

The baseline methods we compare are m-TransH \cite{jf17k}, RAE \cite{rae}, NaLP-Fix \cite{hinge} (an improved version of NaLP \cite{NaLP}), HINGE \cite{hinge}, vanilla Transformer \cite{stare}, and STARE \cite{stare}. For the evaluation metric, we use the filtered setting \cite{transe} for computing mean reciprocal rank (MRR) and hits at 1 and 10 (H@1 and H@10). Higher MRR, H@1, and H@10 scores indicate better
performance. The metrics are computed for head entity and tail entity prediction separately and then averaged.


\begin{table}[!tp]
\caption{Dataset Statistics}
\begin{tabular}{lccc}
\hline
  Dataset           &  \#Entities  & \#Relations  & \#Statements   \\ \hline
 JF17K  & 28,645  & 322      & 100,947   \\
 WikiPeople &  34,839    & 375        &  369,866    \\
 WD50K & 47,156 &  532   & 236,507 \\ \hline
\end{tabular}
\label{tab:data}
\end{table}

\subsection{Main Results}
\label{main}

Table \ref{tab:main} shows the HKG completion results on all the three datasets. We see that our model \Modelshortsp consistently outperforms the strongest baseline method \Baselineshort except slightly lower H@10 on the WikiPeople dataset. Our model gets 7.0\% higher H@1 and 3.6\% higher H@1 on the WikiPeople dataset and WD50K dataset respectively. Moreover, our model outperforms other baselines by a large margin: more than 5.3\%, 13.6\%, and 24.4\% in MRR on the three datasets respectively.

\begin{figure}[!tp]
    \centering
    \includegraphics[width=8.5cm]{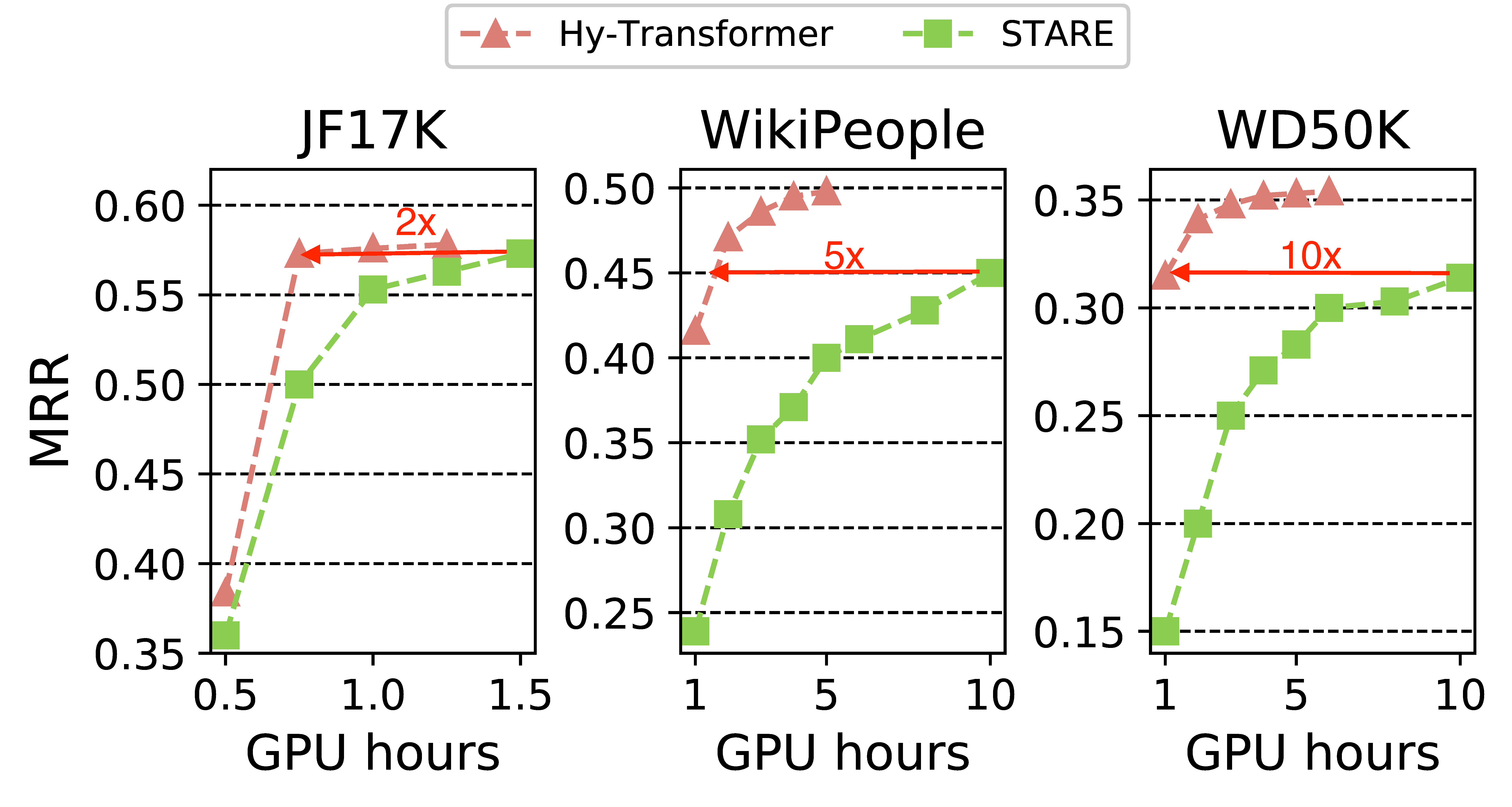}
    \caption{MRR performance (averaged over 5 different runs) of \Modelshortsp and \Baselineshortsp over training time.}
    \label{fig:time}
\end{figure}

We also show the training time comparison between our model and \Baselineshortsp 
in Figure \ref{fig:time}. As shown in the figure, our model takes much less training time to achieve the same test MRR results. For example, in the WikiPeople dataset, our model achieved 0.45 MRR within 2 hours while \Baselineshortsp takes about 10 hours, which is 5 times slower. In the WD50K dataset, our model takes 1 hour to achieve 0.31 MRR while \Baselineshortsp costs about 10 hours, which is 10 times slower. These results demonstrate that our model can be trained much more efficiently than \Baselineshortsp while still achieving on-par or better results. This also shows that the larger the KG is, the more efficiency improvement our model can gain, which is consistent with the analysis in Section \ref{replace}.

\begin{table*}[!tp]
\caption{Experiment results in HKG completion on all the three datasets. The baseline results are directly taken from \cite{stare}. The average results over 5 different runs are reported for our model.}
\setlength{\tabcolsep}{2mm}{
\begin{tabular}{lccccccccc}
\hline
                         & \multicolumn{3}{c}{WikiPeople}                & \multicolumn{3}{c}{JF17K}                  & \multicolumn{3}{c}{WD50K}                              \\ \cmidrule(lr){2-4}\cmidrule(lr){5-7}\cmidrule(lr){8-10} 
\multirow{-2}{*}{Methods} & MRR           & H@1           & H@10          & MRR                       & H@1           & H@10          & MRR           & H@1                        & H@10                       \\ \hline
m-TransH \cite{jf17k}                & -             & 0.063         & 0.300           & - & 0.206         & 0.463         & -             & -                          & -                          \\
RAE  \cite{rae}                    & -             & 0.059         & 0.306         & -                         & 0.215         & 0.469         & -             & -                          & -                          \\
NaLP-Fix \cite{hinge}                 & 0.420          & 0.343         & 0.556         & 0.245                     & 0.185         & 0.358         & 0.177         & 0.131 & 0.264 \\
HINGE   \cite{hinge}                 & 0.476         & 0.415         & 0.585         & 0.449                     & 0.361         & 0.624         & 0.243         & 0.176 & 0.377 \\
Transformer   \cite{stare}           & 0.469         & 0.403         & 0.586         & 0.512                     & 0.434         & 0.665         & 0.286         & 0.222                      & 0.406                      \\
\Baselineshortsp  \cite{stare}       & 0.491         & 0.398         & \textbf{0.648}         & 0.574                     & 0.496         & 0.725         & 0.349        & 0.271                      & 0.496                      \\ 
\Modelshortsp         & \textbf{0.501}  & \textbf{0.426} & 0.634  & \textbf{0.582}            & \textbf{0.501}  & \textbf{0.742} & \textbf{0.356}  & \textbf{0.281}              & \textbf{0.498}           \\ \hline
\end{tabular}
}
\label{tab:main}
\end{table*}

\subsection{Ratio of Statements with Qualifiers}
\label{qualifier}

In this section, we study the model performance under different ratios of hyper-relational statements (statements with qualifiers). In the WD50K dataset, about 13\% of statements contain qualifiers while the remaining 87\% only have main triplets. We further conduct experiments on three different variants of the WD50K dataset provides by \cite{stare}: WD50K (33), WD50K (66), and WD50K (100), which contain approximately 33\%, 66\%,  and 100\% of hyper-relational statements, respectively. For example, on WD50K (100) dataset, all the statements contain one or more qualifiers.

The results are shown in Table \ref{fig:qualifier}, where we see that our model outperforms \Baselineshortsp and Transformer even without training with the auxiliary task proposed in Section \ref{aux}. By the auxiliary training task, we see that the MRR results of our model improve by 3.3\%, 5.1\%, and 5.7\% on WD50K (33), WD50K (66), and WD50K (100) respectively. This demonstrates the effectiveness of our proposed training task. Moreover, the improvement becomes larger as the ratio of hyper-relational statements increases, which makes sense because the augmented training signals comes from statements with qualifiers. 


\begin{table}[!tp]
\caption{MRR results in HKG completion on three variants of the WD50K dataset with different qualifier ratios. The baseline results are directly taken from \cite{stare}. The averaged results over 5 different runs are reported for our model. HT w/o aux stands for our model Hy-Transformer without the auxiliary training task.}
\begin{tabular}{l|ccc}
\hline
Method         & WD50K (33) & WD50K (66) & WD50K (100) \\ \hline
Transformer \cite{stare}    & 0.276 & 0.404 & 0.562      \\
\Baselineshortsp \cite{stare}          & 0.331 & 0.481 & 0.654     \\
HT w/o aux & 0.332 & 0.490 & 0.661 \\
Hy-Transformer & \textbf{0.343} & \textbf{0.515} & \textbf{0.699}      \\  \hline
\end{tabular}
\label{fig:qualifier}
\end{table}


\subsection{Ablation Study}
\label{ablation}

In this section, we illustrate the effectiveness of each component for embedding processing in \Modelshortsp by conducting ablation study on the WD50K (100) dataset mentioned in Section \ref{qualifier}. 

As shown in Table \ref{tab:ablation},
without entity embedding layer normalization, the performance of our model drops significantly by 6.6\% of MRR. This shows that controlling the scale of entity embeddings is very important in this task.  Entity embedding dropout also plays an essential role by improving 4.0\% of MRR since it helps to prevent over-fitting when the number of entity embedding parameters is very large. Layer normalization on relation embeddings can also slightly improve the performance. 
Overall, all the three embedding processing techniques can improve the HKG completion performance.

\begin{table}[!tp]
\caption{HKG completion results on the WD50K (100) datasets. LN is short for Layer Normalization.}
\begin{tabular}{l|ccc}
\hline
Method           & MRR   & H@1   & H@10  \\ \hline
\Modelshort     & 0.699 & 0.637 & 0.812 \\
\ \ w/o entity embedding LN      & 0.653  & 0.572 & 0.803 \\
\ \ w/o entity embedding dropout & 0.672 & 0.617 & 0.772 \\
\ \ w/o relation embedding LN    & 0.693  & 0.632  & 0.806  \\
\hline
\end{tabular}
\label{tab:ablation}
\end{table}

\section{Conclusion}


We tackle the problem of hyper-relational KG completion. Based on the current best-performing model \Baselineshort, we propose two novel strategies to improve both its efficiency and effectiveness: (1) replacing the computation-heavy GNN module with light-weight techniques, layer normalization and dropout, on entity/relation embeddings; (2) adding an auxiliary training task which predicts the missing qualifier entity for boosting the prediction power on HKG completion. Our model \Modelshortsp consistently outperforms all the baseline methods over three benchmark datasets and is significantly more efficient than \Baselineshort. In the future, we plan to apply our model into larger scale of HKGs like the full Wikidata with millions of entities and extend our model to incorporate additional context information like the text descriptions of entities and relations.







\bibliographystyle{ACM-Reference-Format}
\bibliography{sample-base}


\begin{thebibliography}{20}


\ifx \showCODEN    \undefined \def \showCODEN     #1{\unskip}     \fi
\ifx \showDOI      \undefined \def \showDOI       #1{#1}\fi
\ifx \showISBNx    \undefined \def \showISBNx     #1{\unskip}     \fi
\ifx \showISBNxiii \undefined \def \showISBNxiii  #1{\unskip}     \fi
\ifx \showISSN     \undefined \def \showISSN      #1{\unskip}     \fi
\ifx \showLCCN     \undefined \def \showLCCN      #1{\unskip}     \fi
\ifx \shownote     \undefined \def \shownote      #1{#1}          \fi
\ifx \showarticletitle \undefined \def \showarticletitle #1{#1}   \fi
\ifx \showURL      \undefined \def \showURL       {\relax}        \fi
\providecommand\bibfield[2]{#2}
\providecommand\bibinfo[2]{#2}
\providecommand\natexlab[1]{#1}
\providecommand\showeprint[2][]{arXiv:#2}

\bibitem[\protect\citeauthoryear{Ba, Kiros, and Hinton}{Ba
  et~al\mbox{.}}{2016}]%
        {layernorm}
\bibfield{author}{\bibinfo{person}{Jimmy~Lei Ba}, \bibinfo{person}{Jamie~Ryan
  Kiros}, {and} \bibinfo{person}{Geoffrey~E Hinton}.}
  \bibinfo{year}{2016}\natexlab{}.
\newblock \showarticletitle{Layer normalization}.
\newblock \bibinfo{journal}{\emph{arXiv preprint arXiv:1607.06450}}
  (\bibinfo{year}{2016}).
\newblock


\bibitem[\protect\citeauthoryear{Bollacker, Evans, Paritosh, Sturge, and
  Taylor}{Bollacker et~al\mbox{.}}{2008}]%
        {bollacker2008freebase}
\bibfield{author}{\bibinfo{person}{Kurt Bollacker}, \bibinfo{person}{Colin
  Evans}, \bibinfo{person}{Praveen Paritosh}, \bibinfo{person}{Tim Sturge},
  {and} \bibinfo{person}{Jamie Taylor}.} \bibinfo{year}{2008}\natexlab{}.
\newblock \showarticletitle{Freebase: a collaboratively created graph database
  for structuring human knowledge}. In \bibinfo{booktitle}{\emph{Proceedings of
  the 2008 ACM SIGMOD international conference on Management of data}}.
  \bibinfo{pages}{1247--1250}.
\newblock


\bibitem[\protect\citeauthoryear{Bordes, Usunier, Garc{\'{\i}}a{-}Dur{\'{a}}n,
  Weston, and Yakhnenko}{Bordes et~al\mbox{.}}{2013}]%
        {transe}
\bibfield{author}{\bibinfo{person}{Antoine Bordes}, \bibinfo{person}{Nicolas
  Usunier}, \bibinfo{person}{Alberto Garc{\'{\i}}a{-}Dur{\'{a}}n},
  \bibinfo{person}{Jason Weston}, {and} \bibinfo{person}{Oksana Yakhnenko}.}
  \bibinfo{year}{2013}\natexlab{}.
\newblock \showarticletitle{Translating Embeddings for Modeling
  Multi-relational Data}. In \bibinfo{booktitle}{\emph{Advances in Neural
  Information Processing Systems 26: 27th Annual Conference on Neural
  Information Processing Systems 2013. Proceedings of a meeting held December
  5-8, 2013, Lake Tahoe, Nevada, United States}},
  \bibfield{editor}{\bibinfo{person}{Christopher J.~C. Burges},
  \bibinfo{person}{L{\'{e}}on Bottou}, \bibinfo{person}{Zoubin Ghahramani},
  {and} \bibinfo{person}{Kilian~Q. Weinberger}} (Eds.).
  \bibinfo{pages}{2787--2795}.
\newblock
\urldef\tempurl%
\url{https://proceedings.neurips.cc/paper/2013/hash/1cecc7a77928ca8133fa24680a88d2f9-Abstract.html}
\showURL{%
\tempurl}


\bibitem[\protect\citeauthoryear{Chen, Chen, Hui, Wu, Li, and Lin}{Chen
  et~al\mbox{.}}{2020}]%
        {chen2020knowledge}
\bibfield{author}{\bibinfo{person}{Riquan Chen}, \bibinfo{person}{Tianshui
  Chen}, \bibinfo{person}{Xiaolu Hui}, \bibinfo{person}{Hefeng Wu},
  \bibinfo{person}{Guanbin Li}, {and} \bibinfo{person}{Liang Lin}.}
  \bibinfo{year}{2020}\natexlab{}.
\newblock \showarticletitle{Knowledge Graph Transfer Network for Few-Shot
  Recognition}. In \bibinfo{booktitle}{\emph{The Thirty-Fourth {AAAI}
  Conference on Artificial Intelligence, {AAAI} 2020, The Thirty-Second
  Innovative Applications of Artificial Intelligence Conference, {IAAI} 2020,
  The Tenth {AAAI} Symposium on Educational Advances in Artificial
  Intelligence, {EAAI} 2020, New York, NY, USA, February 7-12, 2020}}.
  \bibinfo{publisher}{{AAAI} Press}, \bibinfo{pages}{10575--10582}.
\newblock
\urldef\tempurl%
\url{https://aaai.org/ojs/index.php/AAAI/article/view/6630}
\showURL{%
\tempurl}


\bibitem[\protect\citeauthoryear{Dettmers, Minervini, Stenetorp, and
  Riedel}{Dettmers et~al\mbox{.}}{2018}]%
        {conve}
\bibfield{author}{\bibinfo{person}{Tim Dettmers}, \bibinfo{person}{Pasquale
  Minervini}, \bibinfo{person}{Pontus Stenetorp}, {and}
  \bibinfo{person}{Sebastian Riedel}.} \bibinfo{year}{2018}\natexlab{}.
\newblock \showarticletitle{Convolutional 2D Knowledge Graph Embeddings}. In
  \bibinfo{booktitle}{\emph{Proceedings of the Thirty-Second {AAAI} Conference
  on Artificial Intelligence, (AAAI-18), the 30th innovative Applications of
  Artificial Intelligence (IAAI-18), and the 8th {AAAI} Symposium on
  Educational Advances in Artificial Intelligence (EAAI-18), New Orleans,
  Louisiana, USA, February 2-7, 2018}},
  \bibfield{editor}{\bibinfo{person}{Sheila~A. McIlraith} {and}
  \bibinfo{person}{Kilian~Q. Weinberger}} (Eds.). \bibinfo{publisher}{{AAAI}
  Press}, \bibinfo{pages}{1811--1818}.
\newblock
\urldef\tempurl%
\url{https://www.aaai.org/ocs/index.php/AAAI/AAAI18/paper/view/17366}
\showURL{%
\tempurl}


\bibitem[\protect\citeauthoryear{Devlin, Chang, Lee, and Toutanova}{Devlin
  et~al\mbox{.}}{2019}]%
        {devlin2018bert}
\bibfield{author}{\bibinfo{person}{Jacob Devlin}, \bibinfo{person}{Ming-Wei
  Chang}, \bibinfo{person}{Kenton Lee}, {and} \bibinfo{person}{Kristina
  Toutanova}.} \bibinfo{year}{2019}\natexlab{}.
\newblock \showarticletitle{{BERT}: Pre-training of Deep Bidirectional
  Transformers for Language Understanding}. In
  \bibinfo{booktitle}{\emph{Proceedings of the 2019 Conference of the North
  {A}merican Chapter of the Association for Computational Linguistics: Human
  Language Technologies, Volume 1 (Long and Short Papers)}}.
  \bibinfo{publisher}{Association for Computational Linguistics},
  \bibinfo{address}{Minneapolis, Minnesota}, \bibinfo{pages}{4171--4186}.
\newblock
\urldef\tempurl%
\url{https://doi.org/10.18653/v1/N19-1423}
\showDOI{\tempurl}


\bibitem[\protect\citeauthoryear{Fatemi, Taslakian, V{\'{a}}zquez, and
  Poole}{Fatemi et~al\mbox{.}}{2020}]%
        {hype}
\bibfield{author}{\bibinfo{person}{Bahare Fatemi}, \bibinfo{person}{Perouz
  Taslakian}, \bibinfo{person}{David V{\'{a}}zquez}, {and}
  \bibinfo{person}{David Poole}.} \bibinfo{year}{2020}\natexlab{}.
\newblock \showarticletitle{Knowledge Hypergraphs: Prediction Beyond Binary
  Relations}. In \bibinfo{booktitle}{\emph{Proceedings of the Twenty-Ninth
  International Joint Conference on Artificial Intelligence, {IJCAI} 2020}},
  \bibfield{editor}{\bibinfo{person}{Christian Bessiere}} (Ed.).
  \bibinfo{publisher}{ijcai.org}, \bibinfo{pages}{2191--2197}.
\newblock
\urldef\tempurl%
\url{https://doi.org/10.24963/ijcai.2020/303}
\showDOI{\tempurl}


\bibitem[\protect\citeauthoryear{Galkin, Trivedi, Maheshwari, Usbeck, and
  Lehmann}{Galkin et~al\mbox{.}}{2020}]%
        {stare}
\bibfield{author}{\bibinfo{person}{Mikhail Galkin}, \bibinfo{person}{Priyansh
  Trivedi}, \bibinfo{person}{Gaurav Maheshwari}, \bibinfo{person}{Ricardo
  Usbeck}, {and} \bibinfo{person}{Jens Lehmann}.}
  \bibinfo{year}{2020}\natexlab{}.
\newblock \showarticletitle{Message Passing for Hyper-Relational Knowledge
  Graphs}. In \bibinfo{booktitle}{\emph{Proceedings of the 2020 Conference on
  Empirical Methods in Natural Language Processing (EMNLP)}}.
  \bibinfo{publisher}{Association for Computational Linguistics},
  \bibinfo{address}{Online}, \bibinfo{pages}{7346--7359}.
\newblock
\urldef\tempurl%
\url{https://doi.org/10.18653/v1/2020.emnlp-main.596}
\showDOI{\tempurl}


\bibitem[\protect\citeauthoryear{Guan, Jin, Wang, and Cheng}{Guan
  et~al\mbox{.}}{2019}]%
        {NaLP}
\bibfield{author}{\bibinfo{person}{Saiping Guan}, \bibinfo{person}{Xiaolong
  Jin}, \bibinfo{person}{Yuanzhuo Wang}, {and} \bibinfo{person}{Xueqi Cheng}.}
  \bibinfo{year}{2019}\natexlab{}.
\newblock \showarticletitle{Link Prediction on N-ary Relational Data}. In
  \bibinfo{booktitle}{\emph{The World Wide Web Conference, {WWW} 2019, San
  Francisco, CA, USA, May 13-17, 2019}},
  \bibfield{editor}{\bibinfo{person}{Ling Liu}, \bibinfo{person}{Ryen~W.
  White}, \bibinfo{person}{Amin Mantrach}, \bibinfo{person}{Fabrizio
  Silvestri}, \bibinfo{person}{Julian~J. McAuley}, \bibinfo{person}{Ricardo
  Baeza{-}Yates}, {and} \bibinfo{person}{Leila Zia}} (Eds.).
  \bibinfo{publisher}{{ACM}}, \bibinfo{pages}{583--593}.
\newblock
\urldef\tempurl%
\url{https://doi.org/10.1145/3308558.3313414}
\showDOI{\tempurl}


\bibitem[\protect\citeauthoryear{Guo, Zhuang, Qin, Zhu, Xie, Xiong, and He}{Guo
  et~al\mbox{.}}{2020}]%
        {kgrecsurvey}
\bibfield{author}{\bibinfo{person}{Qingyu Guo}, \bibinfo{person}{Fuzhen
  Zhuang}, \bibinfo{person}{Chuan Qin}, \bibinfo{person}{Hengshu Zhu},
  \bibinfo{person}{Xing Xie}, \bibinfo{person}{Hui Xiong}, {and}
  \bibinfo{person}{Qing He}.} \bibinfo{year}{2020}\natexlab{}.
\newblock \showarticletitle{A survey on knowledge graph-based recommender
  systems}.
\newblock \bibinfo{journal}{\emph{IEEE Transactions on Knowledge and Data
  Engineering}} (\bibinfo{year}{2020}).
\newblock


\bibitem[\protect\citeauthoryear{He, Balakrishnan, Eric, and Liang}{He
  et~al\mbox{.}}{2017}]%
        {dialogue}
\bibfield{author}{\bibinfo{person}{He He}, \bibinfo{person}{Anusha
  Balakrishnan}, \bibinfo{person}{Mihail Eric}, {and} \bibinfo{person}{Percy
  Liang}.} \bibinfo{year}{2017}\natexlab{}.
\newblock \showarticletitle{Learning Symmetric Collaborative Dialogue Agents
  with Dynamic Knowledge Graph Embeddings}. In
  \bibinfo{booktitle}{\emph{Proceedings of the 55th Annual Meeting of the
  Association for Computational Linguistics (Volume 1: Long Papers)}}.
  \bibinfo{publisher}{Association for Computational Linguistics},
  \bibinfo{address}{Vancouver, Canada}, \bibinfo{pages}{1766--1776}.
\newblock
\urldef\tempurl%
\url{https://doi.org/10.18653/v1/P17-1162}
\showDOI{\tempurl}


\bibitem[\protect\citeauthoryear{Ji, Pan, Cambria, Marttinen, and Yu}{Ji
  et~al\mbox{.}}{2020}]%
        {kgsurvey}
\bibfield{author}{\bibinfo{person}{Shaoxiong Ji}, \bibinfo{person}{Shirui Pan},
  \bibinfo{person}{Erik Cambria}, \bibinfo{person}{Pekka Marttinen}, {and}
  \bibinfo{person}{Philip~S Yu}.} \bibinfo{year}{2020}\natexlab{}.
\newblock \showarticletitle{A survey on knowledge graphs: Representation,
  acquisition and applications}.
\newblock \bibinfo{journal}{\emph{arXiv preprint arXiv:2002.00388}}
  (\bibinfo{year}{2020}).
\newblock


\bibitem[\protect\citeauthoryear{Kingma and Ba}{Kingma and Ba}{2015}]%
        {adam}
\bibfield{author}{\bibinfo{person}{Diederik~P. Kingma} {and}
  \bibinfo{person}{Jimmy Ba}.} \bibinfo{year}{2015}\natexlab{}.
\newblock \showarticletitle{Adam: {A} Method for Stochastic Optimization}. In
  \bibinfo{booktitle}{\emph{3rd International Conference on Learning
  Representations, {ICLR} 2015, San Diego, CA, USA, May 7-9, 2015, Conference
  Track Proceedings}}, \bibfield{editor}{\bibinfo{person}{Yoshua Bengio} {and}
  \bibinfo{person}{Yann LeCun}} (Eds.).
\newblock
\urldef\tempurl%
\url{http://arxiv.org/abs/1412.6980}
\showURL{%
\tempurl}


\bibitem[\protect\citeauthoryear{Lin, Chen, Chen, and Ren}{Lin
  et~al\mbox{.}}{2019}]%
        {lin2019kagnet}
\bibfield{author}{\bibinfo{person}{Bill~Yuchen Lin}, \bibinfo{person}{Xinyue
  Chen}, \bibinfo{person}{Jamin Chen}, {and} \bibinfo{person}{Xiang Ren}.}
  \bibinfo{year}{2019}\natexlab{}.
\newblock \showarticletitle{{K}ag{N}et: Knowledge-Aware Graph Networks for
  Commonsense Reasoning}. In \bibinfo{booktitle}{\emph{Proceedings of the 2019
  Conference on Empirical Methods in Natural Language Processing and the 9th
  International Joint Conference on Natural Language Processing
  (EMNLP-IJCNLP)}}. \bibinfo{publisher}{Association for Computational
  Linguistics}, \bibinfo{address}{Hong Kong, China},
  \bibinfo{pages}{2829--2839}.
\newblock
\urldef\tempurl%
\url{https://doi.org/10.18653/v1/D19-1282}
\showDOI{\tempurl}


\bibitem[\protect\citeauthoryear{Liu, Yao, and Li}{Liu et~al\mbox{.}}{2020}]%
        {liu2020generalizing}
\bibfield{author}{\bibinfo{person}{Yu Liu}, \bibinfo{person}{Quanming Yao},
  {and} \bibinfo{person}{Yong Li}.} \bibinfo{year}{2020}\natexlab{}.
\newblock \showarticletitle{Generalizing Tensor Decomposition for N-ary
  Relational Knowledge Bases}. In \bibinfo{booktitle}{\emph{{WWW} '20: The Web
  Conference 2020, Taipei, Taiwan, April 20-24, 2020}},
  \bibfield{editor}{\bibinfo{person}{Yennun Huang}, \bibinfo{person}{Irwin
  King}, \bibinfo{person}{Tie{-}Yan Liu}, {and} \bibinfo{person}{Maarten van
  Steen}} (Eds.). \bibinfo{publisher}{{ACM} / {IW3C2}},
  \bibinfo{pages}{1104--1114}.
\newblock
\urldef\tempurl%
\url{https://doi.org/10.1145/3366423.3380188}
\showDOI{\tempurl}


\bibitem[\protect\citeauthoryear{Rosso, Yang, and Cudr{\'{e}}{-}Mauroux}{Rosso
  et~al\mbox{.}}{2020}]%
        {hinge}
\bibfield{author}{\bibinfo{person}{Paolo Rosso}, \bibinfo{person}{Dingqi Yang},
  {and} \bibinfo{person}{Philippe Cudr{\'{e}}{-}Mauroux}.}
  \bibinfo{year}{2020}\natexlab{}.
\newblock \showarticletitle{Beyond Triplets: Hyper-Relational Knowledge Graph
  Embedding for Link Prediction}. In \bibinfo{booktitle}{\emph{{WWW} '20: The
  Web Conference 2020, Taipei, Taiwan, April 20-24, 2020}},
  \bibfield{editor}{\bibinfo{person}{Yennun Huang}, \bibinfo{person}{Irwin
  King}, \bibinfo{person}{Tie{-}Yan Liu}, {and} \bibinfo{person}{Maarten van
  Steen}} (Eds.). \bibinfo{publisher}{{ACM} / {IW3C2}},
  \bibinfo{pages}{1885--1896}.
\newblock
\urldef\tempurl%
\url{https://doi.org/10.1145/3366423.3380257}
\showDOI{\tempurl}


\bibitem[\protect\citeauthoryear{Srivastava, Hinton, Krizhevsky, Sutskever, and
  Salakhutdinov}{Srivastava et~al\mbox{.}}{2014}]%
        {srivastava2014dropout}
\bibfield{author}{\bibinfo{person}{Nitish Srivastava},
  \bibinfo{person}{Geoffrey Hinton}, \bibinfo{person}{Alex Krizhevsky},
  \bibinfo{person}{Ilya Sutskever}, {and} \bibinfo{person}{Ruslan
  Salakhutdinov}.} \bibinfo{year}{2014}\natexlab{}.
\newblock \showarticletitle{Dropout: a simple way to prevent neural networks
  from overfitting}.
\newblock \bibinfo{journal}{\emph{The journal of machine learning research}}
  \bibinfo{volume}{15}, \bibinfo{number}{1} (\bibinfo{year}{2014}),
  \bibinfo{pages}{1929--1958}.
\newblock


\bibitem[\protect\citeauthoryear{Vaswani, Shazeer, Parmar, Uszkoreit, Jones,
  Gomez, Kaiser, and Polosukhin}{Vaswani et~al\mbox{.}}{2017}]%
        {transformer}
\bibfield{author}{\bibinfo{person}{Ashish Vaswani}, \bibinfo{person}{Noam
  Shazeer}, \bibinfo{person}{Niki Parmar}, \bibinfo{person}{Jakob Uszkoreit},
  \bibinfo{person}{Llion Jones}, \bibinfo{person}{Aidan~N. Gomez},
  \bibinfo{person}{Lukasz Kaiser}, {and} \bibinfo{person}{Illia Polosukhin}.}
  \bibinfo{year}{2017}\natexlab{}.
\newblock \showarticletitle{Attention is All you Need}. In
  \bibinfo{booktitle}{\emph{Advances in Neural Information Processing Systems
  30: Annual Conference on Neural Information Processing Systems 2017, December
  4-9, 2017, Long Beach, CA, {USA}}},
  \bibfield{editor}{\bibinfo{person}{Isabelle Guyon}, \bibinfo{person}{Ulrike
  von Luxburg}, \bibinfo{person}{Samy Bengio}, \bibinfo{person}{Hanna~M.
  Wallach}, \bibinfo{person}{Rob Fergus}, \bibinfo{person}{S.~V.~N.
  Vishwanathan}, {and} \bibinfo{person}{Roman Garnett}} (Eds.).
  \bibinfo{pages}{5998--6008}.
\newblock
\urldef\tempurl%
\url{https://proceedings.neurips.cc/paper/2017/hash/3f5ee243547dee91fbd053c1c4a845aa-Abstract.html}
\showURL{%
\tempurl}


\bibitem[\protect\citeauthoryear{Wen, Li, Mao, Chen, and Zhang}{Wen
  et~al\mbox{.}}{2016}]%
        {jf17k}
\bibfield{author}{\bibinfo{person}{Jianfeng Wen}, \bibinfo{person}{Jianxin Li},
  \bibinfo{person}{Yongyi Mao}, \bibinfo{person}{Shini Chen}, {and}
  \bibinfo{person}{Richong Zhang}.} \bibinfo{year}{2016}\natexlab{}.
\newblock \showarticletitle{On the Representation and Embedding of Knowledge
  Bases beyond Binary Relations}. In \bibinfo{booktitle}{\emph{Proceedings of
  the Twenty-Fifth International Joint Conference on Artificial Intelligence,
  {IJCAI} 2016, New York, NY, USA, 9-15 July 2016}},
  \bibfield{editor}{\bibinfo{person}{Subbarao Kambhampati}} (Ed.).
  \bibinfo{publisher}{{IJCAI/AAAI} Press}, \bibinfo{pages}{1300--1307}.
\newblock
\urldef\tempurl%
\url{http://www.ijcai.org/Abstract/16/188}
\showURL{%
\tempurl}


\bibitem[\protect\citeauthoryear{Zhang, Li, Mei, and Mao}{Zhang
  et~al\mbox{.}}{2018}]%
        {rae}
\bibfield{author}{\bibinfo{person}{Richong Zhang}, \bibinfo{person}{Junpeng
  Li}, \bibinfo{person}{Jiajie Mei}, {and} \bibinfo{person}{Yongyi Mao}.}
  \bibinfo{year}{2018}\natexlab{}.
\newblock \showarticletitle{Scalable Instance Reconstruction in Knowledge Bases
  via Relatedness Affiliated Embedding}. In
  \bibinfo{booktitle}{\emph{Proceedings of the 2018 World Wide Web Conference
  on World Wide Web, {WWW} 2018, Lyon, France, April 23-27, 2018}},
  \bibfield{editor}{\bibinfo{person}{Pierre{-}Antoine Champin},
  \bibinfo{person}{Fabien~L. Gandon}, \bibinfo{person}{Mounia Lalmas}, {and}
  \bibinfo{person}{Panagiotis~G. Ipeirotis}} (Eds.).
  \bibinfo{publisher}{{ACM}}, \bibinfo{pages}{1185--1194}.
\newblock
\urldef\tempurl%
\url{https://doi.org/10.1145/3178876.3186017}
\showDOI{\tempurl}


\end{thebibliography}


\end{document}